\documentclass{article}

\usepackage{PRIMEarxiv}

\usepackage[utf8]{inputenc} % allow utf-8 input
\usepackage[T1]{fontenc}    % use 8-bit T1 fonts
\usepackage{hyperref}       % hyperlinks
\usepackage{url}            % simple URL typesetting
\usepackage{booktabs}       % professional-quality tables
\usepackage{amsfonts}       % blackboard math symbols
\usepackage{nicefrac}       % compact symbols for 1/2, etc.
\usepackage{microtype}      % microtypography
\usepackage{lipsum}
\usepackage{fancyhdr}       % header
\usepackage{graphicx}       % graphics
\usepackage{subfig}
\usepackage{cite}
\usepackage{times}
\usepackage{epsfig}
\usepackage{graphicx}
\usepackage{amsmath}
\usepackage{amssymb}
\usepackage{float}
\usepackage{multirow}
\usepackage{caption}
\graphicspath{ {./Figure/} }
\title{AugShuffleNet: communicate more, compute less}

\author{
  LongQing Ye
}

\begin{document}
\maketitle

\begin{abstract}
%Orthogonal to network width and depth, this paper shows the proportion of short paths in the neural network could be another crucial design dimension for efficient model design. 

As a remarkable compact model, ShuffleNetV2 offers a good example to design efficient ConvNets but its limit is rarely noticed. In this paper, we rethink the design pattern of ShuffleNetV2 and find that the channel-wise redundancy problem still constrains the efficiency improvement of Shuffle block in the wider ShuffleNetV2. To resolve this issue, we propose another augmented variant of shuffle block in the form of bottleneck-like structure and more implicit short connections. To verify the effectiveness of this building block, we further build a more powerful and efficient model family, termed as AugShuffleNets. Evaluated on the CIFAR-10 and CIFAR-100 datasets, AugShuffleNet consistently outperforms ShuffleNetV2 in terms of accuracy with less computational cost and fewer parameter count.
\end{abstract}

\section{Introduction}
Since the success of AlexNet \cite{alex}, convolutional neural networks (CNNs, or ConvNets) have become dominant in various vision tasks including image classification, object detection and semantic segmentation. Typical models including VGG, GoogLeNet and Resnet \cite{vgg,googlenet,resnet} achieved remarkable performance. However, the accuracy improvement usually involves scaling up network in the form of more layers or more channels per layer, further leading to an increasing amount of computational cost and parameters. In recent years, many real-time CNN-based applications are deployed on resource-constrained platforms such as sensors and smart phones, requiring CNN models to be computationally efficient and of rapid response. Therefore, how to design efficient models has become an important research topic.

Many efforts has been dedicated to improve the efficiency of ConvNets. Currently, there are two main representative and complementary schemes: model compression and compact model design. The former kind aims to reduce the redundancy of a larger model without significant degeneration in performance via network pruning, quantization, low-rank, etc. However, the upper limit of model performance tends to be determined by the pre-trained network. The latter kind is trying to build compact models trained from scratch following . Typical instances like MobileNetV1-V3 and ShuffleNetV1-V2 utilize efficient operators, building block or network architecture search algorithms, providing significant insights for compact model design.

In this paper, we revisit the design of building block in ShuffleNetV2 and design a more efficient and powerful CNN model named AugShuffleNet

%two crucial variables split ratio and communication frequency into ShuffleNetV2, a novel paradigm to trade off between efficiency and performance is proposed to design efficient models. 

%To construct deep neural networks, most of models would use residual connection. Essentially, residual connection provides short paths for optimization during the back-propagation process, allowing parameters at the shallow layers to be updated and thus alleviating the performance degradation due to overlong optimization paths. The high efficiency of ShuffleNetV2 partially comes from its smart strategy to introduce implicit short paths via information communication between two branches, which could avoid element-wise operation and reduce considerable memory access cost.

%Inspired by this, we postulate that the proportion of short and long paths could be a crucial factor to influence the model performance.

%we notice that the influence of cross-layer information communication is not well explored.

%further design a more efficient and powerful 

%more intermediate representations in the network are fully utilized.

%and introduce two crucial variables split ratio and communication frequency
%to 

%Following such framework,  and modify it into a more efficient and powerful model named []. 

%our approach allows by trading off between the frequency of cross-layer information and computational cost of transformation layers.

%In summary, the contributions of this paper are list:

\section{Related Work}

\subsection{Compact Model Design}
Composition of low-rank ($1\times3$, $3\times1$) filters \cite{low-rank} or sparse filters \cite{shufflenet} are proposed to approximate dense convolution filter. Those work indicates that less redundant filters can bring a great reduction of FLOPs and parameters while maintaining the performance of models. Group convolution has been viewed as a standard operator in modern compact models \cite{resnext,shufflenet}. Depth-wise convolution is an extreme form of group convolution, in which each channel presents a group. ShuffleNetV1 \cite{shufflenet} uses group convolution to replace $1\times1$ convolution and further introduces the operation of "Channel Shuffle" to improve cross-group information communication. MobilenetV1 \cite{mobilenetv1} utilizes depth-wise convolution and pointwise convolution to construct a lightweight model for mobile platforms. ResNet\cite{resnet} and MobileNetV2 \cite{mobilenetv2} adopt a bottleneck structure to alleviate the burden of heavy computation for channel expansion. MobileNetV3 and other work \cite{mobilenetv3,once,darts} introduce the neural architecture search algorithms into compact model design, significantly reducing the cost of manual design.

\subsection{Model Compression}

Comlementary to compact model design, model compression is another approach to further shrink pre-trained models. Network prunning \cite{channelprun} removes redundant and non-informative connections or channels. Model quantization \cite{quantized} aims to represent stored weights at a low cost for model compression and calculation acceleration. Knowledge distillation \cite{distilling} transfers refined knowledge from "teacher network" into "student network", simplifying the process of suppressing redundant information. In addition, efficient convolution algorithms like FFT \cite{fft} and winograd \cite{fast} are explored to speed up the implement of convolutional layer without any modification of network design.

\subsection{Short Connection} 
ResNets and Highway Networks \cite{highway} introduce skip connections to allow training deeper neural networks. Stochastic depth \cite{stochastic} reduces training depth of ResNets by randomly skipping layers. SkipNet futher utilizes gating mechanism to learn how to skip layers for both training and inference procedures. Nowadays residual connection proposed by ResNet has been a popular technique to construct deep neural nerworks \cite{transformer,resnext} in vision and natural language processing domain. Research \cite{ensemble} regards ResNet as a collection of many paths of different length and discovers that deep ResNet updates weights mainly via short paths during the training procedure. DenseNet \cite{densenet} introduces densely connections by connecting each layer to every other layer in the networks to increase more short path. In general, under the framework of backpropagation neural network, short connections create more shorter paths to allow cross-layer gradient flow for smoother optimization while long paths in deep neural networks enable more complex information extraction and larger capacity to store temporary information during the dynamics process of training.

\section{Approach}

\begin{figure*}[ht]
\centering
\subfloat[Shuffle block in ShuffleNetV2]{\label{shuffle}\includegraphics[width=0.3\linewidth]{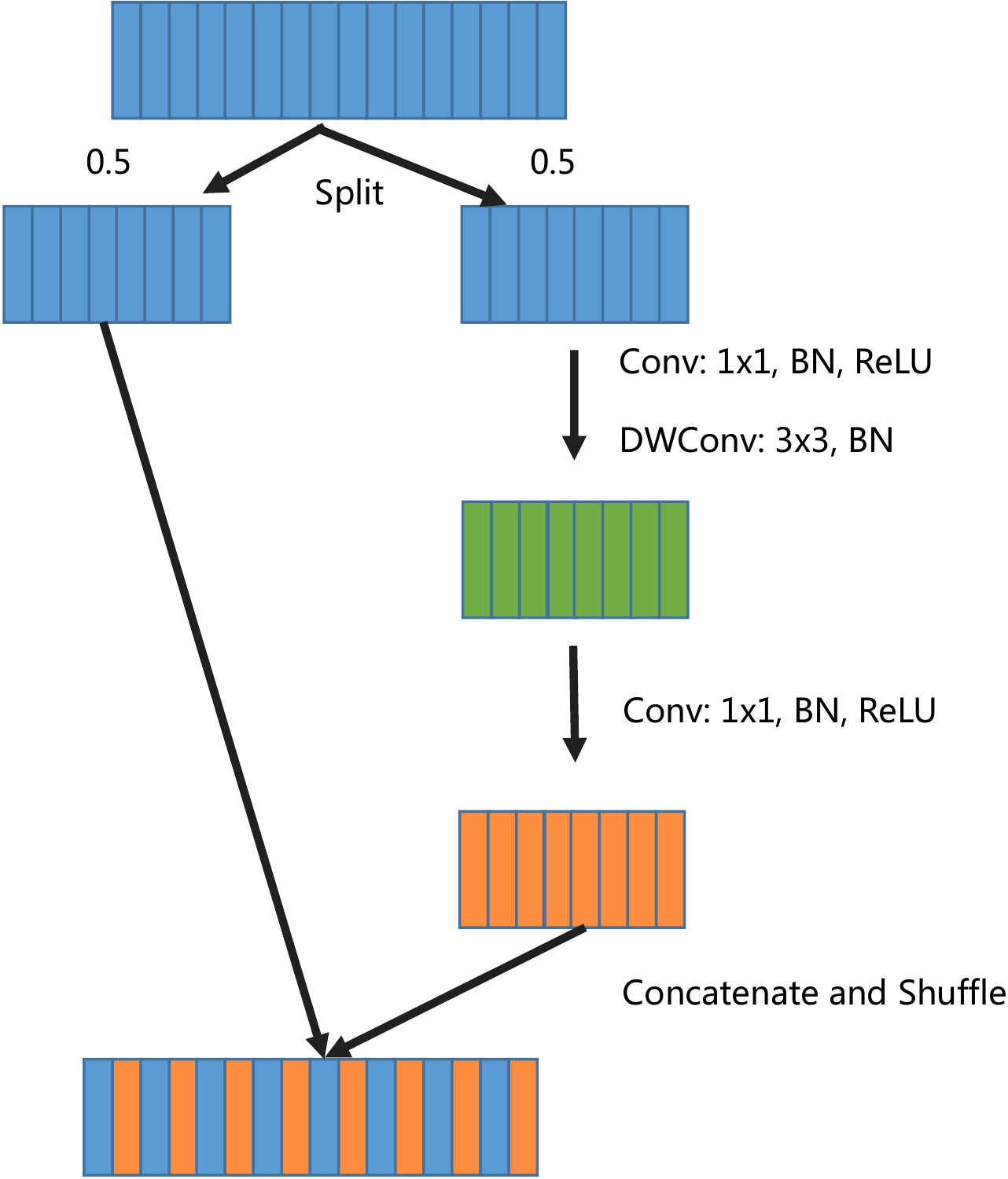}}
\hspace{12mm}
\subfloat[Shuffle block of this paper]{\label{bank}\includegraphics[width=0.3\linewidth]{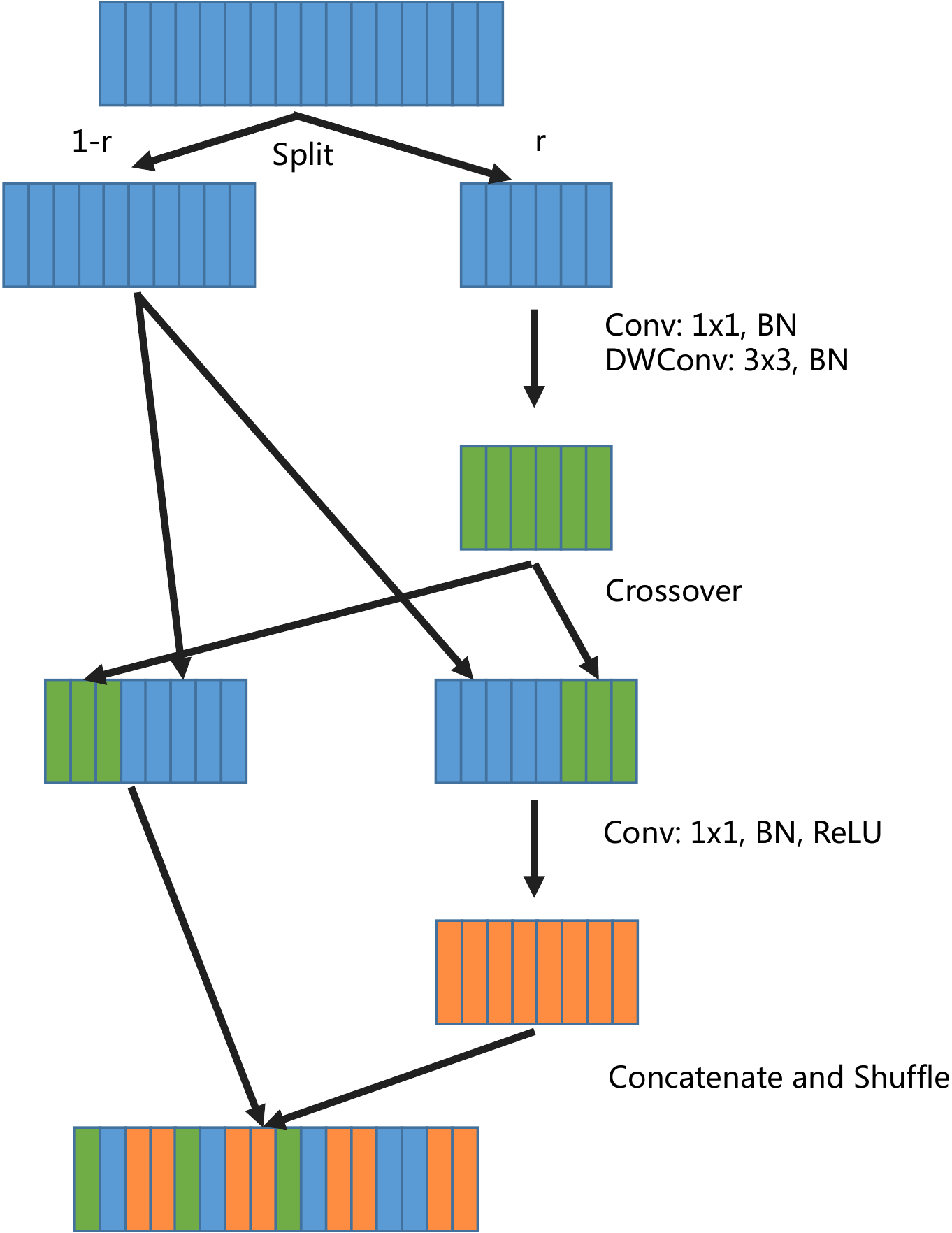}}

\caption{Illustration of building blocks in the channel-wise perspective}
\label{blocks}

\end{figure*}

\subsection{Rethink Shuffle Block in ShuffleNetV2} \label{MPFC}
Essentially, residual connection is a element-wise operation with low computational consumption but high memory access cost. Distinct from common ConvNets, ShuffleNetV2 \cite{shufflenetv2} abandons residual connection to introduce short paths for neural network. Instead, to pursue higher efficiency, it combines "Channel Split" and "Channel Shuffle" to achieve cross-layer information communication, generating a proportion of short paths from the perspective of backpropagation.

Illustrated in Figure \ref{shuffle}, ShuffleNetV2 adopts split-transform-fuse strategy to design basic building block named shuffle block. The whole pipeline of shuffle block can be concluded as three stages:

\noindent\textbf{Split:} 
Channels of feature maps are split into two parts. One part is conserved in memory, another is fed into the next stage for high-level information extraction. Note that the conserved part is named "Feature Bank" in this paper. The number of feature maps in "Feature Bank"  is determined by the parameter split ratio. It is seen that split ratio is a fixed parameter 0.5 in ShuffleNetV2.

\noindent\textbf{Transform:} Transformation stage consists of three layers: $1\times 1$ regular convolution, $3\times 3$ depth-wise convolution and regular $1\times 1$ convolution in order, serving as a learnable "Feature Extractor". 

\noindent\textbf{Fuse:} In this stage, feature maps coming from "Feature Bank" and previous stage are merged by concatenating along channel-wise dimension. Then, those merged channels are rearranged in an interleaving way, which is called "Channel Shuffle". "Channel Shuffle" enables information communication between the two branches.

Essentially, "Feature Bank" can be regarded as a auxiliary memory component where feature maps are stored temporarily. The success of ShuffleNetV2 lies in its use of "Feature Bank" via which to control the information communication across different layers. Here, "Channel Split" and "Channel Shuffle" are two main interactive operations between "Feature Bank" and "Feature Extractor". "Channel Split" determines the proportion of input feature maps which should be stored. "Channel Shuffle" is used to fuse old and new feature maps for the next building block.
As a result, partial channels from shallow layers can be stored in "Feature Bank" and periodically fused into deeper layers of the network, introducing more short paths to replace residual connection \cite{resnet} and achieving a similar effect of feature reuse \cite{densenet}.

\begin{figure}[ht]
\centering
\includegraphics[scale=0.5]{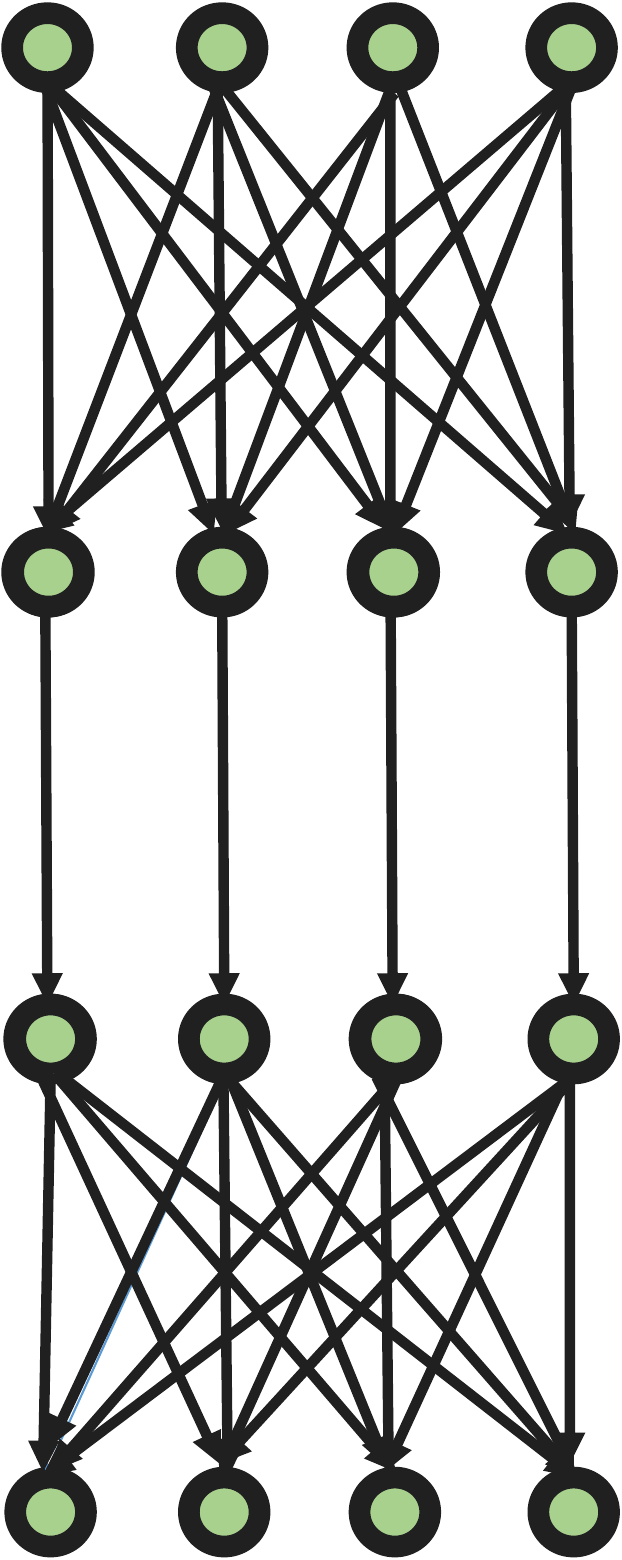}

\caption{}
\label{topo}
\end{figure}

 Benefited from its unique design pattern, ShuffleNetV2 shows expressive balance between performance and efficiency. However, we notice that there still exists remaining room to improve for ShuffleNetV2:
 
\begin{itemize}
    \item [1.] It is seen taht information communication is only limited to the exit of the shuffle block via the operation of "Channel Shuffle". Intermediate information generated by the first and second layer of shuffle block can not be fully exploited by other building blocks.
    
    \item [2.] In the shuffle block of ShuffleNetV2, three convolutional layers are forced to keep the same number of input and output channels. As illustrated in Figure \ref{topo}, they follows fully connected topological pattern in the channel-wise dimension. When network width (the number of channels) increases drastically, shuffle block in ShuffleNetv2 will produce more channel-wise redundancy, which constrains the design space of ShuffleNetV2 for better efficiency.
\end{itemize}

\subsection{Augmented Shuffle Block}

We propose another more powerful and efficient shuffle block illustrated in Figure \ref{bank}. Compared to the original shuffle block shown in \ref{shuffle}, this paper makes following modifications:
\begin{itemize}
    \item [M1.] Split ratio $r$ is set as a variable, which can flexibly adjust the number of channels fed into "Feature Extractor" and further control the efficiency of the whole building block in terms of computational cost, parameter consumption and inference speed. For simplicity,  the first and second layers ($1\times 1$ regular convolution and $3\times 3$ depth-wise convolution) keep the same number of input and output channels.
    
    \item[M2.] We introduce another operation termed "Channel Crossover" to exchange information after the dewpth-wise convolutional layer. "Channel Crossover" deposit partial new feature maps into "Feature Bank" and withdraw more old feature maps to achieve channel-wise expansion without computational cost. By adding interaction mechanism between intermediate layers and "Feature Bank", information is fully utilized.
    
    \item[M3.] We remove ReLU funcntion after the first convolutional layer.
    
\end{itemize}

M1, M2 and M3 work together to improve efficiency and make model gain better representational ability. when $r < 0.5$, the first and second layers in shuffle block would be more efficient than original shuffle block

\subsection{Efficiency Analysis}
Suppose both of shuffle blocks take $M$ feature maps with a spatial size of $D_f \times D_f$ as input. The split ratio is set to $r$, the kernel size of depth-wise convolution is defined as $D_k$. Hence, the computational cost of augmented shuffle block can be formulated as
\begin{equation}
\begin{aligned}
C = r^{2}M^{2}{D_f}^{2}+ rMD_{f}^{2}D_{k}^{2} + \frac{1}{4}M^{2}{D_f}^{2}
\end{aligned}
\label{calculate}
\end{equation}

The computational cost of original shuffle block in ShuffleNetV2 can be also obtained via Equation \ref{calculate}, we only need to set $r$ to 0.5. Therefore, the ratio of computational cost of two shuffle blocks in AugShuffleNet and ShuffleNetV2 respectively is

\begin{equation}
\begin{aligned}
\alpha = \frac{C_{r}}{C_{r=0.5}}=\frac{(4r^{2}+1)M+4rD_{k}^2}{2M+2D_{k}^2} \approx \frac{4r^{2}+1}{2}
\end{aligned}
\label{a1}
\end{equation}

As depth-wise convolution is a quite efficient operator, its overhead can be ignored. Here, we can obtain an approximate expression of $\alpha$. It is seen $\alpha$ is almost only determined by split ratio $r$. It means that split ration $r$ can be used as a global hyper-parameter to reduce the high-dimensional channel-wise redundancy from two $1\times 1$ convolutional layers and adjust the efficiency of AugShuffleNet.

Similarly, the  parameter consumption $P$ of augmented shuffle block in AugShuffleNet can be formulated as
\begin{equation}
\begin{aligned}
P = r^{2}M^{2}+ rMD_{k}^{2} + \frac{1}{4}M^{2}
\end{aligned}
\label{calculate_P}
\end{equation}

the ratio of parameter consumption is also equal to $\alpha$:

\begin{equation}
\begin{aligned}
\beta = \frac{P_{r}}{P_{r=0.5}}=\frac{(4r^{2}+1)M+4rD_{k}^2}{2M+2D_{k}^2} \approx \frac{4r^{2}+1}{2}
\end{aligned}
\label{b}
\end{equation}

In general, as $r$ decreases, augmented shuffle block becomes more efficient compared to original shuffle block in ShuffleNetV2.

\section{Network Design}
The architecture of AugShuffleNet is shown in Table \ref{arch}. We offer three models with different network width ($0.5\times$, $1.0\times$ and $1.5\times$). To ensure fair comparison, we follow the architecture configuration of ShuffleNetV2. The only difference is that we set the number of channels for Stage2, Stage3 and Stage 4 in AugShuffleNet $1.0\times$ to 120, 240 and 480 while ShuffleNetV2 $1.0\times$'s are 116, 232 and 464. As we need proper channel partition to explore the effect of different split ratio. Every stage in Table \ref{arch} includes a down-sampling block ($Stride=2$) and repeated normal blocks ($Stride=1$). In this paper, we adopt original down-sampling block in ShuffleNetV2 and use augmented shuffle block illustrated in Figure \ref{bank} as normal block to build AugShuffleNet. The split ratio $r$ of all of augmented shuffle blocks are forced to keep the same for AugShuffleNet. The default value of $r$ is set to 0.375.

\begin{table}
\centering

\begin{tabular}{|c|c|c|c|c|c|c|c|} 
\hline
\multirow{2}{*}{\begin{tabular}[c]{@{}c@{}}\\Layer\end{tabular}} & \multirow{2}{*}{~Output size} & \multirow{2}{*}{KSize}           & \multirow{2}{*}{Stride} & \multirow{2}{*}{Repeat}     & \multicolumn{3}{c|}{~ Output channels}                              \\ 
\cline{6-8}
                                                                 &                               &                                  &                         &                             & $0.5\times$          & $1\times$            & $1.5\times$           \\ 
\hline
Image                                                            & $32\times 32$                 &                                  &                         &                             & 3                    & 3                    & 3                     \\ 
\hline
\multicolumn{1}{|l|}{~~~~ Conv~~}                                & $32\times 32$                 & \multicolumn{1}{l|}{$3\times 3$} & 1                       & \multicolumn{1}{l|}{~ ~ ~1} & 24                   & 24                   & 24                    \\ 
\hline
\multirow{2}{*}{~ Stage2~~}                                      & $16\times 16$                 & \multirow{2}{*}{$3\times 3$}     & 2                       & 1                           & \multirow{2}{*}{48}  & \multirow{2}{*}{120} & \multirow{2}{*}{176}  \\
                                                                 & $16\times 16$                 &                                  & 1                       & 3                           &                      &                      &                       \\ 
\hline
\multirow{2}{*}{Stage3}                                          & $8\times 8$                   & \multirow{2}{*}{$3\times 3$}     & 2                       & 1                           & \multirow{2}{*}{96}  & \multirow{2}{*}{240} & \multirow{2}{*}{352}  \\
                                                                 & $8\times 8$                   &                                  & 1                       & 7                           &                      &                      &                       \\ 
\hline
\multirow{2}{*}{Stage4}                                          & $4\times 4$                   & \multirow{2}{*}{$3\times 3$}     & 2                       & 1                           & \multirow{2}{*}{192} & \multirow{2}{*}{480} & \multirow{2}{*}{704}  \\
                                                                 & $4\times 4$                   &                                  & 1                       & 3                           &                      &                      &                       \\ 
\hline
Conv                                                             & $4\times 4$                   & $1\times 1$                      & 1                       & 1                           & 1024                 & 1024                 & 1024                  \\ 
\hline
GlobalPool                                                       & $1\times 1$                   & $4\times 4$                      &                         &                             &                      &                      &                       \\ 
\hline
FC                                                               &                               &                                  &                         &                             & num\_classes         & num\_classes         & num\_classes          \\
\hline
\end{tabular}
\caption{Architecture of AugShuffleNet}
\label{arch}
\end{table}

\section{Experiment and Result}

\textbf{Datasets} All models are evaluated on the datasets CIFAR-10 and CIFAR-100. The two CIFAR datasets consist of colored natural images with $32\times32$ pixels. CIFAR-10 and CIFAR-100 consist of images drawn from 10 classes and 100 classes respectively. The training and test sets contain 50,000 and 10,000 images respectively for both two datasets. Images in the training set are augmented by random horizontal flip and random crop (4 pixels are padded on each side, and a 32×32 crop is randomly sampled from the padded image).

 \textbf{Training Settings} All models are trained 300 epochs using cosine learning rate decay for both of CIFAR-10 and CIFAR-100. The initial learning rate is set as 0.1. we adopt Stochastic Gradient Descend (SGD) optimizer (momentum parameter is 0.9, nesterov is set to True) with the batch size of 128. The weight decay is set to 1e-4.

\begin{table}[ht]

\centering
\begin{tabular}{|c|l|l|l|l|l|l|l|l|} 
\hline
\multirow{2}{*}{Model} & \multirow{2}{*}{Params} & \multirow{2}{*}{FLOPs} & \multicolumn{6}{c|}{~ Acc (\%)}                              \\ 
\cline{4-9}
                       &                         &                        & \#run1 & \#run2 & \#run3 & \#run4 & \#run5 & \#Average  \\ 
\hline

ShuffleNetV2  $1.5\times$        &        2.49M                  &         94.27M               &      93.93  &  93.96      &   93.86     & 93.98       &   94.11     &    93.97        \\ 

AugShuffleNet  $1.5\times$        &      2.22M                   &         85.38M               &   \textbf{94.31}     &    \textbf{94.41}    &   \textbf{94.21}     &   \textbf{94.44}     &   \textbf{94.62}     &   \textbf{94.40}         \\ 

\cline{1-1}

ShuffleNetV2  $1.0\times$         &      1.26M                   &    45.01M                    &   93.12     &   93.21     &  93.17      &  93.30      &   93.42     &   93.24         \\ 

AugShuffleNet  $1.0\times$         &     1.21M                    &     43.56M                  &    93.63    &    93.78    &  93.95      &   93.96     &    93.87    &  \textbf{93.84}          \\ 

\cline{1-1}

ShuffleNetV2  $0.5\times$         &      0.35M                   &    10.91M                    &   90.62     &   90.95     &  90.67    &  90.79      &   91.12    &   90.83        \\ 

AugShuffleNet  $0.5\times$         &     0.33M                    &     10.20M                  &    90.85    &    90.87    &  91.21     &   91.39     &    90.92    &  \textbf{91.05}        \\ 

\cline{1-1}

\hline
\end{tabular}
\caption{Comparison of models on CIFAR-10}
\label{c10}

\end{table}

\begin{table}[ht]

\centering
\begin{tabular}{|c|l|l|l|l|l|l|l|l|} 
\hline
\multirow{2}{*}{Model} & \multirow{2}{*}{Params} & \multirow{2}{*}{FLOPs} & \multicolumn{6}{c|}{~ Acc (\%)}                              \\ 
\cline{4-9}
                       &                         &                        & \#run1 & \#run2 & \#run3 & \#run4 & \#run5 & \#Average  \\ 
\hline

ShuffleNetV2    $1.5\times$       &    2.58M                     &     94.36M                   &     74.53   &   74.66     &  74.83      &   74.76     &    74.70    &  74.70          \\ 

AugShuffleNet  $1.5\times$          &    2.32M                  &         85.47M  & \textbf{75.65}       & \textbf{75.75}       & \textbf{75.98}       &  \textbf{76.44}      & \textbf{75.75}       &    \textbf{75.91}        \\ 
\cline{1-1}

ShuffleNetV2  $1.0\times$         &     1.36M                    &    45.10M                    &      73.24        &   73.06     &   72.91     &   72.70     &   73.49   & 73.08     \\ 

AugShuffleNet  $1.0\times$          &     1.30M      &       43.65M                 & 74.07       &  74.10      &    74.31    &  74.45      &   73.88     &    \textbf{74.16}        \\
\cline{1-1}

ShuffleNetV2  $0.5\times$         &     0.44M                    &    11.00M                    &     68.97        &   68.46     &   68.97     &   68.39    &   68.67   & 68.69    \\ 

AugShuffleNet  $0.5\times$          &     0.42M      &       10.29M                 & 69.19       &  69.03      &    69.29    &  68.93   &  69.35     &    \textbf{69.15}       \\
\hline
\end{tabular}
\caption{Comparison of models on CIFAR-100}
\label{c100}

\end{table}

\subsection{Comparison to Other Models}
To verify our approach, we choose ShuffleNetV2 as baselines. Accuracy, computational cost and parameter count are selected as evaluation metrics. Note that, term FLOPs means multiply-adds (MAdd) in this paper. Extensive experiments are conducted on image classification datasets CIFAR-10 and CIFAR-100. The result for every model are obtained via $5$ runs to ensure the reliability of experiments. Result are reported in Table \ref{c10} and Table \ref{c100}. It is clear that our models consistently outperform ShuffleNetV2 model in accuracy aspect on both CIAR-10 and CIFAR-100 with better efficiency in terms of computational cost and parameter count. The results under different network width ($0.5\times$, $1.0\times$ and $1.5\times$) is sufficient to demonstrate the advantage of augmented shuffle block in AugShuffleNets.

\subsection{Ablation Study}
Split ratio $r$ is an important hyper-parameter for AugShuffleNet, which directly determines the efficiency of augmented shuffle block. To explore the impact of split ratio $r$, we compare the performance and efficiency between AugShuffleNets $1.5\times$ with different siplt ratio ($0.125$, $0.25$ and $0.375$)  and ShuffleNetV2 $1.5\times$.

\begin{figure*}[ht]
\centering

\subfloat[Computational cost comparison]{\includegraphics[width=1\linewidth]{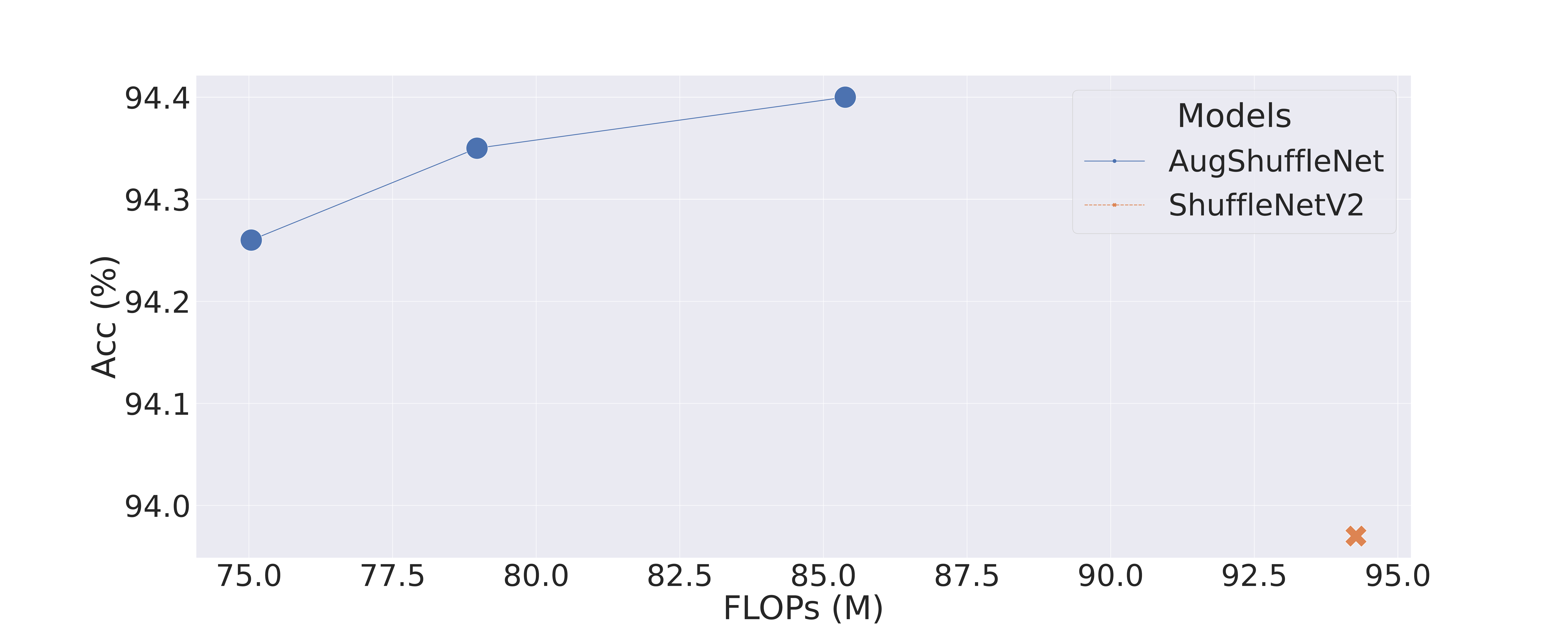}}
\hspace{20mm}
\subfloat[Parameter count comparison]{\includegraphics[width=1\linewidth]{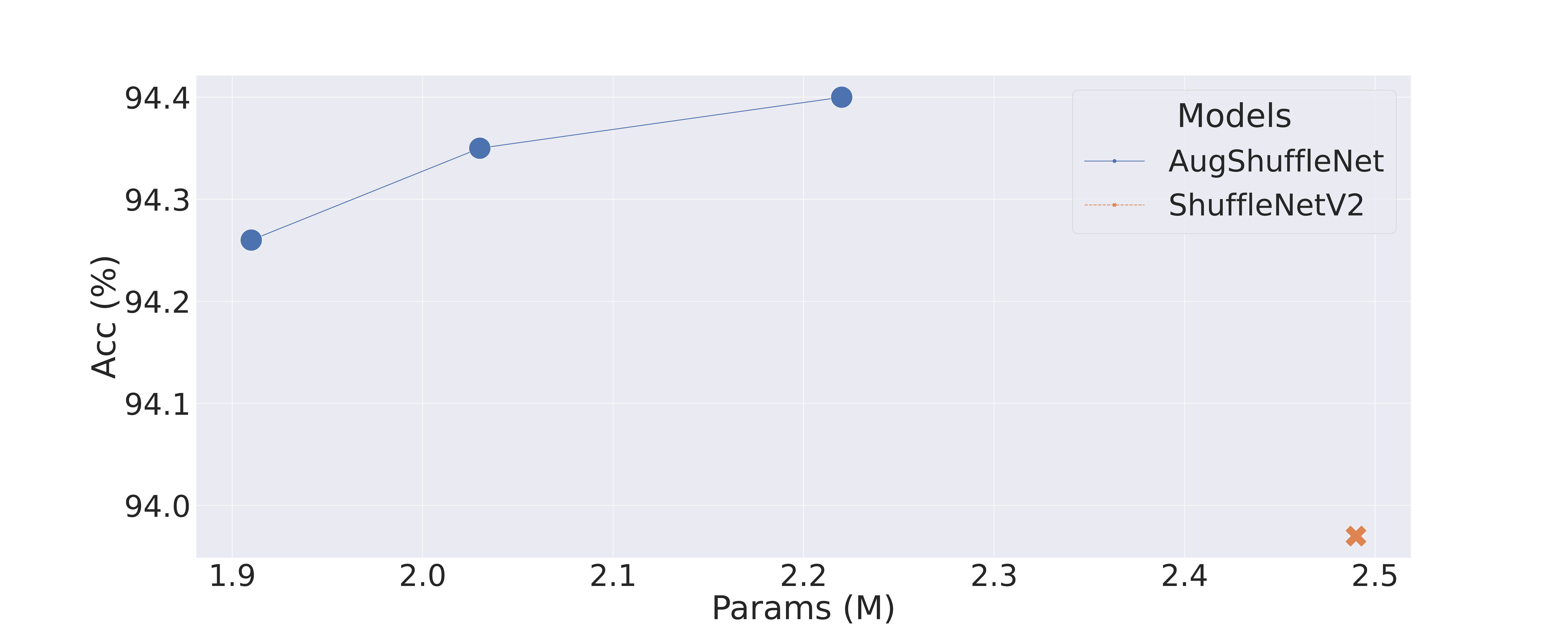}}

\caption{Comparison between AugShuffleNets $1.5\times$ with different split ratio ($0.125$, $0.25$ and $0.375$) and ShuffleNetV2 $1.5\times$ on CIFAR-10.}
\label{s_c10}
\end{figure*}

\begin{figure*}[ht]
\centering
\subfloat[Computational cost comparison]{\includegraphics[width=1\linewidth]{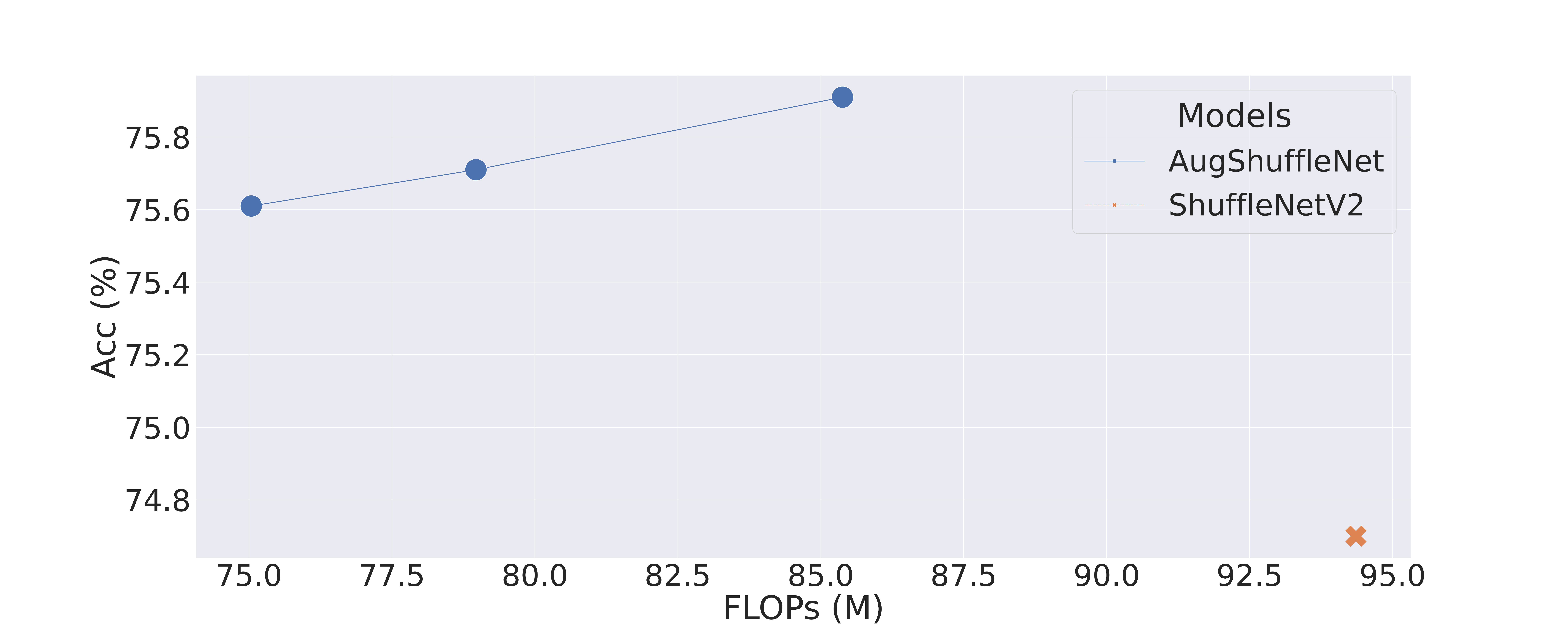}}
\hspace{20mm}
\subfloat[Parameter count comparison]{\includegraphics[width=1\linewidth]{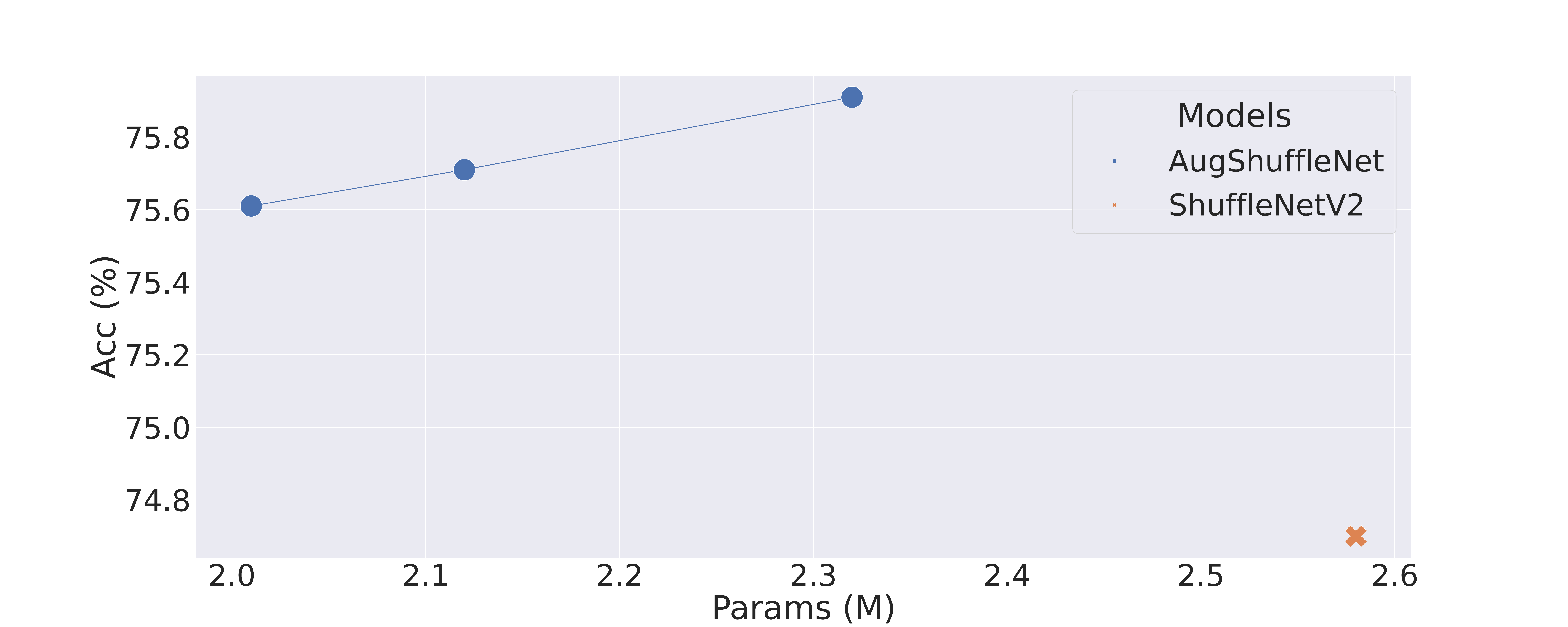}}
\caption{Comparison between AugShuffleNets $1.5\times$ with different split ratio ($0.125$, $0.25$ and $0.375$) and ShuffleNetV2 $1.5\times$ on CIFAR-100.}
\label{s_c100}
\end{figure*}

As illustrated in Figure \ref{s_c10} and Figure \ref{s_c100}, all AugShuffleNet $1.5 \times$ variants obtain higher accuracy than ShuffleNetV2 $1.5\times$ with less computational cost and fewer parameters on both CIFAR-10 and CIFAR-100. It is noteworthy that AugShuffleNet $1.5 \times$ ($r=0.125$) is 0.29\%(0.91\%) more accurate than AugShuffleNet $1.5 \times$ with 79.6\% (79.6\%) computational cost and 76.7\% (77.9\%) parameter count on CIFAR-10 (CIFAR-100).

The above results indicate that the channel-wise redundancy in AugShuffleNet $1.5\times$ is quite considerable for small datasets CIFAR-10 and CIFAR-100. This kind of redundancy problem results from the design pattern of original shuffle block. Our augmented shuffle block is more flexible in adjusting the overhead of the whole building block, finally reducing the channel-wise redundancy due to large network width.

\section{Conclusion}
We propose a more flexible and efficient building block named augmented shuffle block based on ShuffleNetV2. By replacing original shuffle block in ShuffleNetV2, we can see obvious improvement in both performance and efficiency. Our experimental results show that augmented shuffle block is a potential building block to construct compact models for resource-constrained platforms.

\bibliographystyle{unsrt}  
\bibliography{references}

\end{document}